\documentclass{article}
\usepackage[T1]{fontenc} 
\usepackage[utf8]{inputenc} 
\usepackage{ismir,amsmath,cite,url}
\usepackage{graphicx}
\usepackage{color}

 \usepackage[bookmarks=false,hidelinks]{hyperref}

\usepackage{lineno}

\title{Multilingual Music Genre Embeddings for Effective Cross-Lingual Music Item Annotation}

\threeauthors
  {Elena V. Epure}{Deezer Research \\{}} 
  {Guillaume Salha} {Deezer Research \\ {\tt research@deezer.com}}
  {Romain Hennequin}{Deezer Research\\{} }

\def\authorname{E.V. Epure, G. Salha, and R. Hennequin}

\begin{document}

\maketitle
\begin{abstract}
Annotating music items with music genres is crucial for music recommendation and information retrieval, yet challenging given that music genres are subjective concepts. 
Recently, in order to explicitly consider this subjectivity, the annotation of music items was modeled as a translation task:
predict for a music item its music genres within a target vocabulary or taxonomy (tag system) from a set of music genre tags originating from other tag systems. 
However, without a parallel corpus, previous solutions could not handle tag systems in other languages, being limited to the English-language only.
Here, by learning multilingual music genre embeddings, we enable cross-lingual music genre translation without relying on a parallel corpus.
First, we apply compositionality functions on pre-trained word embeddings to represent multi-word tags. 
Second, we adapt the tag representations to the music domain by leveraging multilingual music genres graphs with a modified retrofitting algorithm. 
Experiments show that our method: 1) is effective in translating music genres across tag systems in multiple languages (English, French and Spanish);
2) outperforms the previous baseline in an English-language multi-source translation task.
We publicly release the new multilingual data and code.
\end{abstract}
\section{Introduction}\label{sec:introduction}
Music genres are a key characteristic of music items \cite{Mandel2010,Schedl2017}.
In music streaming services, user profiles and interests can be expressed through music genres, tracks and artists can be grouped in genre-specific collections, and content-based recommender systems frequently exploit music genres as item tags. 
However, music genres are difficult to infer due to their subjective nature. Based on their music preferences, musicological knowledge and culture, people inconsistently associate genres to music items \cite{Craft,Lee,Sordo2008}. 
Thus, annotating music items with genres for providing personalized recommendation and retrieval is challenging.

Acknowledging this subjectivity and the absence of a unique genre definition, recent works \cite{Epure2019,Hennequin2018} framed the music genre annotation as a \textit{translation}.
More precisely, given music items annotated with music genres originating from multiple source tag systems such as folksonomies, editorial vocabularies or taxonomies, the goal was to predict the equivalent music genres within a target tag system.
In a supervised setup, the translation relied on a parallel corpus of music items jointly annotated with music genres from the source and target tag systems \cite{Epure2019,Hennequin2018}.
In an unsupervised setup, when the parallel corpus was unavailable, a solution centered on taxonomy alignment was proposed \cite{Epure2019}. 

However, the translation of music genres between \textit{multilingual} sources remains unaddressed when a parallel corpus is unavailable.
The only past unsupervised solution \cite{Epure2019} relied on heuristics specific to the English language, making its adaptation to multilingual tags a challenge.
Here, we propose to perform the unsupervised cross-lingual translation by leveraging multilingual music genre embeddings.
Also, our method to learn these embeddings could be straightforwardly applied to new languages.

The proposed method is further summarised.
First, by acknowledging the compositional nature of music genres (i.e. the meaning of multi-word music genres can be often derived from the meaning of each word), we learn music genre embeddings by applying compositionality functions to pre-trained word vectors \cite{mikolov2018,pennington2014,grave2018}. 
Moreover, as these pre-trained vectors are often trained on language-specific text, we need to align them across languages  \cite{artetxe2018,mikolov2013exploiting}. 

Second, we fit the obtained music genre embeddings into the music domain.
The embeddings learnt on general-language corpora could sometimes be semantically ambiguous. 
For instance, \emph{house} is closer to \emph{building} than to \emph{music} and \emph{jazz} is more similar to \emph{folk} than to \emph{bebop} in fastText \cite{mikolov2018}.
To tackle this problem, we create a music genre knowledge graph from multilingual DBpedia \cite{Auer_2007} that contains multilingual genres as nodes and exhibits different types of music genre relations through its edges. 
Then, we use \textit{retrofitting} \cite{faruqui2015} to encode the relational knowledge from the semantic graph in the embeddings. 
In this work, we modify the original retrofitting algorithm \cite{faruqui2015} to distinguish between two types of relations: equivalence (e.g. \textit{dnb} and \textit{drum'n'bass}) and other types of relatedness such as sub-genres, derivative genres, fusion genres, stylistic origins. 
Besides, we use retrofitting to learn embeddings for music genres that do not exist in the pretrained embedding vocabulary by exploiting their graph relations (e.g. \textit{ethnotronica} and \textit{chillstep} are not in the pretrained fastText vocabulary).

We evaluate the proposed method in two experiments. 
First, we collect a new parallel corpus of music items annotated with music genres in three languages (English, French and Spanish) and demonstrate the effectiveness of our method for unsupervised cross-lingual music genre translation. 
Second, we show that using the embeddings learnt with our method outperforms the previous baseline \cite{Epure2019} in a music genre translation task between multiple English-language tag systems.


\section{Problem Formulation and \\ Related Work}
\label{sec:mus_genre_trans}
Annotating music items from song lyrics or audio content has concentrated significant research efforts in the music information retrieval community \cite{Oramas2017,Hennequin2018,Coviello2011}.
Most existing works fix a tag system and focus on general music genres like \textit{jazz} or \textit{pop} \cite{Craft,Lee,Sordo2008}. 
Nonetheless, the dissimilarity of music genre tag systems and their use in annotations has been recently put forward for consideration, together with the need to take into account tags with increased granularity \cite{Bogdanov2017, Oramas2017}.
In this direction, two previous works \cite{Epure2019,Hennequin2018} have framed the music genre annotation as a tag translation task between various music genre tag systems.

Specifically, 
given a set $\mathcal{S}$ of \textit{source tag systems}, $S = \cup_{E\in\mathcal{S}}E$ the union of all tags across all source tag systems, $\mathcal{P}$ the partitions of $S$ and $T$ a \textit{target tag system}, the goal is to define a translation scoring  $f: \mathcal{P}(S) \rightarrow {\rm I\!R}^{|T|}$ which estimates a score for each target tag from a set of source tags drawn from $S$.
While in this notation a tag system refers to a set of tags, more general representations such as music genre graphs or taxonomies can also include relations between tags \cite{schreiber2016,Achichi2018,Lisena2018}. 

Hennequin et al. \cite{Hennequin2018} proposed two translation strategies, both relying on the existence of a parallel corpus of music items annotated with music genres.
Epure et al. \cite{Epure2019} addressed also 
the unsupervised case, when such a parallel corpus was absent, and designed a knowledge-based method to learn tag embeddings.
This method relied on multiple building blocks corresponding to tag normalization, the construction of an integrated music genre graph bringing together all source and target tag systems, and a taxonomy alignment algorithm mapping each music genre on DBpedia \cite{Auer_2007} tags.
The DBpedia-related building block yielded music genre vectors quantifying the relatedness of the tag under consideration to each DBpedia music genre.
For translation, considering $\{s_{1}, \dots, s_K\}$ source tags and any target tag $t$, $f$ was computed using cosine similarity:
\begin{equation}
f_t(\{s_{1}, s_{2}, \dots, s_K\}) = \sum_{k=1}^{K}\frac{{\textbf{s}_{k}}^T\textbf{t}}{||\textbf{s}_{k}||_2 ||\textbf{t}||_2},
\label{eq:transfinit}
\end{equation}
where $\textbf{s}_{k}$ and $\textbf{t}$ are the vectors corresponding to each $s_{k}$, respectively $t$ and  $|| \cdot ||_2$ is the Euclidian L2-norm.

In the previous unsupervised work, Epure et al. \cite{Epure2019} focused on English-language music genres, claiming that the extension of the knowledge-based method to include multilingual tag systems was feasible since it relied on multilingual DBpedia.
While we agree that it is feasible, the extent to which the introduced method could be easily changed to support other languages is questionable.
Both normalizing tags and mapping music genres into the DBpedia space rely on language-specific heuristics.
For instance, in normalization, heuristics referring to the length of tokens were used.
However, the average word length, hence what is considered as a short or medium-length token depends on the language \cite{smith2012distinct}. 
Then, mapping music genres on English DBpedia genres is limiting because some tags may exist in two languages but not in the English DBpedia.
Computing directly the degree of relatedness of a source tag to a target tag could be a better alternative.

\section{A Multi-Step Method for Learning Music Genre Embeddings}
\label{sec:proposal}
In this work, we propose a method to learn multilingual music genre embeddings that can be easily extended to new languages and support cross-lingual translation.
The first step is to deduce initial embeddings for multi-word music genres by leveraging pre-trained multilingual word embeddings (Section \ref{sec:step1}).
However, directly using these music genre embeddings in cross-lingual translation is prone to under-perform because:
\begin{itemize}
    \item the embeddings often correspond to the most common word senses (e.g. \textit{country} can refer to \textit{nations} or \textit{rock} could be closer in meaning to \textit{stone}) and they are not disambiguated against the music domain.
    \item some music genres could contain rare words which are absent from the pre-trained model vocabulary, resulting in potentially unknown tag embeddings.
\end{itemize}

To address these issues, we complement distributional concept representations with semantics from knowledge bases that expose concept relations. 
Thus, in a second step, we assemble a multilingual music genre graph (Section \ref{sec:step2}).
Then, we adjust the initial tag embeddings to encode the tag relations from the collected graph, ensuring the domain adaptation.
For this, but also to learn embeddings for concepts with unknown vocabulary words, we use \textit{retrofitting} \cite{faruqui2015}, which we modify to reflect the different types of music genre relations (Section \ref{sec:step3}).

\subsection{Initializing Music Genre Embeddings}
\label{sec:step1}
Under the music genre translation framework introduced in Section \ref{sec:mus_genre_trans}, the main goal boils down to quantifying the degree of relatedness of two textual tags. 
This task is widely popular in the natural language processing (NLP) community and contemporary approaches resort to expressing the relatedness as distance between corresponding word embeddings \cite{mikolov2018,pennington2014,grave2018}. 
The mapping of words on embeddings is guided by the distributional hypothesis \cite{harris1954}, which states that words in similar contexts are likely to have similar meanings.
Word embeddings have been proven effective in capturing word syntactic and semantic similarities and in improving downstream NLP tasks such as natural language understanding \cite{lijurafsky2015} and information retrieval \cite{Gysel_2018}.

In order to measure the relatedness of multilingual words using embeddings learnt from monolingual corpora, an alignment between the language-specific embedding spaces is required. 
Through the alignment \cite{joulin2018}, we ensure that multilingual word embeddings are projected into a common space where they are comparable.
Practically, a mapping function between two monolingual word embedding spaces is learnt, for instance by using a bilingual lexicon \cite{mikolov2013exploiting}. 
Effective alignments have been also found using orthogonal Procrustes \cite{joulin2018, artetxe2018}.

Starting from multilingual word vectors, we discuss strategies to initialize the music genre embeddings.
Music genres can contain multiple words.
We claim that the compositionality principle, stating that the meaning of a multi-word expression is dictated by the meaning of each word, often holds for our case.
For instance, \textit{Dance pop} is related to \textit{dance} and \textit{pop} or \textit{Balada romántica} is a type of \textit{ballad} which is \textit{romantic}\footnote{Exceptions from the principle also exist (e.g. \textit{hard rock}).}.
The contemporary approach for compositional embeddings is to learn a function which derives the embeddings of a multi-word expression from the embeddings of its words \cite{shwartz-2019-systematic}.
The function is learnt by minimizing the distance for each multi-word expression between its distributional embedding and its embedding computed from its word embeddings.
Obtaining the distributional embedding for multi-word music genres would be however challenging because sufficiently large corpora with all tags in multiple languages are required.

For this reason, the first music genre initialization strategy we propose consists of a simple compositionality function such as averaging word embeddings (\textit{avg}).
Let $V = \{c_1, c_2, ..., c_n\}$ be the multilingual vocabulary, $c_i$ being a concept composed of at least one word. 
We aim to compute $\hat{\textbf{Q}}\in {\rm I\!R}^{n \times d}$, the embedding matrix for the vocabulary $V$, where $\hat{\textbf{q}}_i \in {\rm I\!R}^d$ denotes the embedding of concept $c_i$.
If $c_i$ is composed of the following words, $\{w_1, w_2, \dots, w_M\}$, $\hat{\textbf{q}}_i$ can be computed as $\frac{1}{M}\sum_{m=1}^M{\textbf{w}_m}$, where $\textbf{w}_m$ is the embedding of the word $w_m$.
Of note is that if $c_i$ contains words absent from the pretrained word embedding vocabulary, the $d$-dimensional null vector, $\textbf{0}_d$, is used as a default.

The second music genre initialization strategy we propose exploits the fact that some words in a compounded expression may be more illustrative than others.
The more frequently a word is observed in a corpus, the more likely it is that the word is common for a language and semantically less informative (e.g. \textit{music} in \textit{post industrial music}). 
Thus, the compositional embedding computation of a multi-word expression can be modified such that the contribution of each word embedding is inversely proportional to its frequency.
Pre-trained word embeddings are generally released sorted by decreasing word corpus frequency. 
Let $z_{w_m}$ be the rank of $w_m$ in this vocabulary.
Then, based on the Mandelbrot's generalization \cite{mandelbrot1953informational} of the Zipf's law \cite{Zipf1949}, its frequency $f_{w_m}$ can be estimated to $f_{w_
{m}} = 1 / (z_{w_{m}} + 2.7)$.

Further, we rely on the smooth inverse frequency (\textit{sif}) based averaging proposed by Arora et al. \cite{arora2016simple} to compute the multi-word expression embeddings.
This method is aligned with our previous observations and proven highly effective compared to more complex neural network-based models on a large diversity of NLP tasks \cite{arora2016simple}.
Given $f_{w_{m}}$ the estimated frequency of the word $w_m$ and $a$ a fixed hyper-parameter\footnote{Experimentally, it has been shown that $a=10^{-3}$ is a suitable choice when using different types of pre-trained word embeddings \cite{arora2016simple}.}, $\hat{\textbf{q}}_i$ is computed as:
\begin{equation}
  \overline{\textbf{q}}_i = \frac{1}{M}\sum_{m=1}^M{\frac{a}{a + f_{w_{m}}}\textbf{w}_{m}}  
\label{eq:sif1}
\end{equation}
\begin{equation}
    \hat{\textbf{q}}_i = \overline{\textbf{q}}_i - \textbf{u}\textbf{u}^T\overline{\textbf{q}}_i
\end{equation}
where $\textbf{u}$ is the first singular vector from the singular value decomposition of $\overline{\textbf{Q}}$ obtained with the Equation \ref{eq:sif1} \cite{Golub:1970}.

\subsection{Assembling a Multilingual Music Genre Graph}
\label{sec:step2}
Previous related work \cite{Epure2019} created a music genre graph by integrating multiple English-language music genre tag systems and a crawled sub-graph of DBpedia through a node English-language based normalization step.
The other music genre tag systems were Lastfm, Tagtraum and Discogs, used in the 2018 MediaEval AcousticBrainz Genre Task \cite{Bogdanov2017}. 
Here, we bypass the language-specific heuristics normalization and propose a more robust alternative.
We crawl a multilingual DBpedia music genre sub-graph and use its words as basis for normalizing new tag systems.

We further detail how we assemble the DBpedia-based music genres graph.
We set the seeds for crawling from: 1) DBpedia entities of type  \textit{MusicGenre}, 2) the music genres of the multilingual DBpedia-based music item corpus (described in Section \ref{sec:datasets}), 3) synonyms of the music genres of the previous two sources, linked through the \textit{wikiPageRedirects} relation.  
We discover new potential music genres by crawling DBpedia entities linked to the seeds through one of the relations: \textit{wikiPageRedirects}, \textit{stylisticOrigin}, \textit{musicSubgenre}, \textit{derivative} and \textit{musicFusionGenre}\footnote{These relation names correspond to the English-language DBpedia. They have often translated versions in DBpedia in other languages.}.
During crawling, seeds are updated with discovered entities that were not visited before, and the crawling goes on until no seeds are left.
This is applied for each language.
Finally, all music genres discovered as yet are connected to their equivalent tags in other languages, when possible (the DBpedia relation \textit{sameAs}).
In a post-processing step, we remove music genre nodes written as free-style text, which do not have DBpedia pages, and the connected components which do not contain at least one high-confidence music genre (empirically, we noticed that the highest-confidence tags were those from the music item corpus).

To ensure that tag systems with different music genre spellings benefit from the multilingual graph, we define a normalization which we apply to both the graph nodes and new tags.
First, we tokenize each tag by non-alphanumeric characters.
Further, as in \cite{Epure2019}, we create prefix trees to split multi-word tags such as \textit{sludgemetal} or \textit{indierock}.
Nevertheless, we do not necessarily aim at a grammatically correct split, but at one based on lemmatized DBpedia music genre words\footnote{Through lemmatization, we retrieve the base form of inflected words using spacy (\url{https://spacy.io}).
Most genre words are generally in their base form (e.g. \textit{jazz}).
However, some other words benefit from this (e.g. \textit{Northern} / \textit{North} or \textit{children} / \textit{child})}.
Namely, if \textit{sludgemetal} is already among the DBpedia music genre words, then there is no further split and we expect its initial embedding to be corrected though the embeddings of its graph neighbors as explained in the next section (Section \ref{sec:step3}).

\subsection{Retrofitting Music Genre Embeddings}
\label{sec:step3}
Retrofitting \cite{faruqui2015} has been proposed as a post-processing step to improve concept embeddings by leveraging existing semantic lexicons or knowledge graphs (e.g. WordNet \cite{Miller_1995}).
More precisely, concept embeddings are modified to also encode concept relations \cite{faruqui2015,lengerich2018,Fang2019,scheepers2018}.

Let $G = (V, E)$ be the graph capturing the semantic relations between the nodes in $V$ through a set of edges $E \subseteq V \times V$.
The objective of retrofitting is to learn $\textbf{Q} \in {\rm I\!R}^{n \times d}$, the new concept embeddings, such that each new embedding $\textbf{q}_i \in {\rm I\!R}^d$ does not stray too far from the initial distributional embedding $\hat{\textbf{q}}_i$, but also becomes closer to the new embeddings of the neighbour vertices $\textbf{q}_j \in {\rm I\!R}^d, j: (i, j) \in E$.
The objective function to minimize is then \cite{faruqui2015}:
\begin{equation}
    \Phi(\textbf{Q}) = \sum_{i\in V}\big( \alpha_i ||\textbf{q}_i - \hat{\textbf{q}}_i||^2_2 + \sum_{j: (i, j)\in E}{\beta_{ij}||\textbf{q}_i - \textbf{q}_j||^2_2} \big)
    \label{eq:objective}
\end{equation}
where $\alpha_i$ and $\beta_{ij}$ are positive scalars specifying the importance given to each component, the initial embedding and each graph neighbor.
As $\Phi$ is convex with respect to $\textbf{Q}$, a solution minimizing the objective function $\Phi$ is found in \cite{faruqui2015} via an iterative strategy derived from Jacobi iteration algorithm \cite{saad2003iterative} that converges for such graph-based propagation problem \cite{saad2003iterative,bengio2006label}. More precisely, until convergence, $\textbf{q}_i$ is iteratively updated as follows:
\begin{equation}
    \textbf{q}_i \xleftarrow{} \frac{\sum_{j:(i,j) \in E}{(\beta_{ij}+ \beta_{ji})\textbf{q}_j} + \alpha_i \hat{\textbf{q}}_i}{\sum_{j:(i,j) \in E}{(\beta_{ij}+\beta_{ji})} + \alpha_i}
    \label{eq:update}
\end{equation}
where $\textbf{Q}$ is initialized to $\hat{\textbf{Q}}$. 
Equation \eqref{eq:update} is not the same as the original one \cite{faruqui2015}.
We observed that, when applying the Jacobi method to optimize equation \eqref{eq:objective}, the contributing terms in the partial derivative with respect to the node $i$ are those where $i$ appears as source as well as target node in the inner sum, leading to a different update.
In \cite{bengio2006label,saha2016dis}, the same conclusion referring to a corrected update rule, different from the initial proposal, is reached.

Faruqui et al. \cite{faruqui2015} set $\alpha_i = 1$ and $\beta_{ij} = \frac{1}{\text{degree}(i)}$ for $(i,j) \in E$, where $\text{degree}(i)$ is the number of neighbors $i$ has in the graph $G$, or $0$ for $(i,j) \not \in E$. 
This choice was largely adopted in other related works \cite{hayes2019,lengerich2018}. 
Speer and Chin \cite{speer2016ensemble} proposed to use a modified version of retrofitting to learn embeddings for unknown vocabulary concepts which are present in the knowledge graph.
For this case, $\alpha_i$ is set to 0 for all unknown vocabulary concepts.
This results in $\textbf{q}_i$ being updated by averaging the embeddings of its neighbours at each iteration.
Despite the change in the update rule we made, compared to the original work, we retain this choice of hyper-parameters as being a reasonable default, and defer the investigation of a more principled way to pick $\alpha_{i}$ and $\beta_{ij}$ to future work.

We further modify retrofitting to take advantage of the different types of music genre relations.
On one hand, music genres can be semantically equivalent to other music genres (the relation types \textit{wikiPageRedirects} and \textit{sameAs}).
On the other hand, music genres can be related to other music genres, but not semantically equivalent (e.g. \textit{stylisticOrigin}).
The change we propose for computing these new embeddings ($\textbf{Q}_{\overline{\beta}}$) is through the coefficients $\beta_{ij}$, making them dependent on music genre relation types:
\begin{equation}
    \overline{\beta}_{ij} = \begin{cases}
    \begin{aligned}
    1 &\quad\text{if } (i,j) \in E_\epsilon \subset E\\
     \beta_{ij}  &\quad\text{if } (i,j) \in E - E_\epsilon\\ 
    0 & \quad\text{otherwise}
\end{aligned}
       \end{cases}
\end{equation}
where $E_\epsilon$ contains edges which represent equivalence relations (\textit{wikiPageRedirects}, \textit{sameAs}); 
$E - E_\epsilon$ contains edges with the remaining relation types (\textit{stylisticOrigin}, \textit{musicSubgenre}, \textit{derivative}, \textit{musicFusionGenre}).

\section{Experiments}
\label{sec:evaluation}
We evaluate the effectiveness of the learnt music genre embeddings, first, in a new cross-lingual music genre translation scenario (Section \ref{sec:results1}) and, second, in an existing English-language multi-source music genre translation task \cite{Bogdanov2017,Epure2019} (Section \ref{sec:results2}).
The languages we focus on for the cross-lingual annotation are English (\textbf{En}), French (\textbf{Fr}) and Spanish (\textbf{Es}).
We start by presenting the parallel corpora used in the experiments (Section \ref{sec:datasets}).
Then, we discuss the detailed evaluation setup (Section \ref{sec:evalsetup}).
The results show that our music genre vectors are highly effective for cross-lingual translation and lead to improved results on the past unsupervised Enligh-language translation task \cite{Epure2019}.

\subsection{Datasets}
\label{sec:datasets}

\begin{table*}
\centering
\begin{tabular}{|l|l|l|l|l|}
    \hline
\textbf{Title} & \textbf{Type} & \textbf{En} & \textbf{Fr} & \textbf{Es} \\ \hline
Morning View & Album & Alternative\_metal, Funk\_rock & Rock\_alternatif &	Metal\_alternativo\\
& & Alternative\_rock, Post-grunge & & Rock\_experimental \\ \hline
Jimi Hendrix & Artist & Hard\_rock, Psychedelic\_rock & Rock\_psychédélique & Blues\_rock, Rock\_psicodélico \\
 &  & Blues, Rhythm\_and\_blues & Blues\_rock, Hard\_rock & Hard\_rock \\ \hline
 Julio Iglesias & Artist & Dance-pop, Latin\_music & Pop\_française	& Pop\_latino, Balada\_romántica \\
  & & Adult\_contemporary\_music &		& Soft\_rock, Adult\_contemporary
\\ \hline
\end{tabular}
\caption{Examples of DBpedia music items annotated with music genres. The tag choices are inconsistent across sources. 
Tags may be adapted to a language (e.g. \textit{Pop\_latino} in \textbf{es}) or may keep the same form in all languages (e.g. \textit{Hard\_rock}).}
\label{tab:examples}
\end{table*}

\begin{table*}
\centering
\begin{tabular}{|c|ccc|ccc|}
    \hline
 & \textbf{En} & \textbf{Fr} & \textbf{Es} & \textbf{Dc} & \textbf{Lf} & \textbf{Tt}  \\
\hline
Music items (tracks, albums, artists) & 48 146 & 30 611 & 34 918 & 1 098 336 & 686 978 & 589 583\\
Unique music genres & 489 & 338 & 491 & 315 & 327 & 296 \\
\hline
\end{tabular}
\caption{Number of music items and unique music genres in the multilingual and the English-language parallel corpora.}
\label{tab:datasets}
\end{table*}

For the cross-lingual translation experiment, we relied on DBpedia \cite{Auer_2007} to collect a parallel corpus.
During an initial manual analysis, we noticed that DBpedia music artists or works could have associated quite different music genres across languages.
We present a few examples in Table \ref{tab:examples}.
Also, when the tags used in annotations were equivalent, they were sometimes partially translated (e.g. \textit{Rock\_alternatif} in \textbf{Fr}), while other times they maintained the same form as in English (e.g. \textit{Soft\_rock} in \textbf{Es}).
We collected DBpedia entities of type \textit{MusicalArtist}, \textit{Band}, or \textit{MusicalWork} with music genres associated in at least two languages.
Then, in a post-processing step, we filtered out the music items with tags that appeared less than $16$ times. 

For the English-language multi-source translation, we use an existing dataset \cite{Bogdanov2017,Epure2019}, which contains tracks annotated with English-language music genres from three sources. 
Discogs (\textbf{Dc}) tags are provided by editors per album, and automatically propagated to each track \cite{Bogdanov2017}.
Lastfm (\textbf{Lf}) and Tagtraum (\textbf{Tt}) tags are created by Internet users per track.
We show in Table \ref{tab:datasets} the number of music items and unique music genres in the new multilingual and the past English-language multi-source parallel corpora.

\subsection{Evaluation Setup}
\label{sec:evalsetup}
We evaluated our models by translating music genre tags associated with tracks from multiple source tag systems to a target tag system.
The translation scoring function computes a score for each tag of the target tag system as the degree of relatedness of the target tag to the input set of source tags.
Compared to Equation \ref{eq:transfinit}, the translation scoring function we use here averages the cosine similarities between each source and target tag embeddings:
\begin{equation}
\hat{f_t}(\{s_{1}, s_{2}, \dots, s_K\}) = \frac{1}{K} f_t(\{s_{1}, s_{2}, \dots, s_K\})
\label{eq:transf}
\end{equation} 

Like in other multi-label prediction tasks \cite{Oramas2017,Ibrahim2020}, we use a ranking metric in evaluation, namely the area under the receiver operating characteristic curve (AUC).
We macro-average the scores and report their mean and standard deviations computed over $4$ folds. 
We split the multi-label data in a stratified way, balancing the overall number of music items and tag distribution across the folds \cite{Sechidis:2011}.

For each experiment, English-language multi-source and cross-lingual, we have three input tag systems represented as partially aligned music genre graphs.
For the multi-source translation, we assemble a graph from the English-language DBpedia music genre sub-graph and the input taxonomies, Discogs, Lastfm and Tagtraum.
For the cross-lingual translation, the new music genre graph, which was assembled as described in Section \ref{sec:step2}, has $10 748$ tags in \textbf{En}, $2 905$ in \textbf{Fr} and $3 988$ in \textbf{Es}.
The translation is performed using annotations from combinations of two out of three tag systems to the kept-out tag system.
We also retain in evaluation annotations which are only from one of the two selected source tag systems. 

In each experiment, we compare the \textit{avg} and \textit{sif} strategies to initialize the music genre vectors.
We report results when using directly the initial embeddings ($\hat{\textbf{Q}}$) in translation or retrofitted with the original method ($\textbf{Q}$) or with our modified version ($\textbf{Q}_{\overline{\beta}}$).
As for the choice of pre-trained word embeddings, we use multilingual fastText \cite{grave2018} which we align with the method proposed by Joulin et al. \cite{joulin2017}.

\subsection{Results on Cross-Lingual Genre Translation}
\label{sec:results1}
In Table \ref{tab:multilang}, we present the results of the cross-lingual music genre translation.
The baseline we propose estimates the relatedness of two tags to be the length of their shortest path in the multilingual DBpedia-based music genre graph.
As a reminder, this graph is partially aligned, meaning that some music genres have equivalent tags in other languages.
As shown in Table \ref{tab:multilang} in parentheses, the baseline scores are quite high proving that the graph is fairly effective for cross-lingual translation on this dataset.
Even so, we are able to exceed these scores by a large margin with our music genre embeddings, initialized with \textit{sif} and retrofitted to take into account the music genre relations.

\begin{table*}
    \centering
    \begin{tabular}{|l|c|c|c|c|c|c|c|}
    \hline
 & \textbf{Baseline}  &  $\hat{\textbf{Q}}$ (\textit{avg}) &$\textbf{Q}$ (\textit{avg}) & $\textbf{Q}_{\overline{\beta}}$ (\textit{avg}) & $\hat{\textbf{Q}}$ (\textit{sif}) &$\textbf{Q}$ (\textit{sif}) & $\textbf{Q}_{\overline{\beta}}$ (\textit{sif}) \\ \hline
   \textbf{En} + \textbf{Es} $\Longrightarrow$ \textbf{Fr} & $85.4 \pm 0.4$ & $73.7 \pm 0.2$ & $77.5 \pm 0.1$ & $ 87.0 \pm 0.2$ & $85.9 \pm 0.1$ & $87.7 \pm 0.1$ & $\textbf{92.3} \pm \textbf{0.1}$  \\ \hline
   \textbf{En} + \textbf{Fr} $\Longrightarrow$ \textbf{Es} & $84.3 \pm 0.2$ & $73.6 \pm 0.3$ & $76.1 \pm 0.2$ & $84.9 \pm 0.2$ & $85.6 \pm 0.0$ & $86.3 \pm 0.1$ & $\textbf{91.3} \pm \textbf{0.1}$ \\ \hline
   \textbf{Fr} + \textbf{Es} $\Longrightarrow$ \textbf{En} & $80.4 \pm 0.1$ & $76.7 \pm 0.4$ & $84.2 \pm 0.2$ & $87.0 \pm 0.3$ & $84.5 \pm 0.2$ & $88.4 \pm 0.3$ & $\textbf{90.2} \pm \textbf{0.2}$  \\ \hline
    \end{tabular}
\caption{Macro-AUC (\%) in cross-lingual music genre translation with standard deviation computed over 4 folds. Results are shown for different embedding initialization (\textit{avg} and \textit{sif}), used directly ($\hat{\textbf{Q}}$) or retrofitted with the original retrofitting ($\textbf{Q}$) or with our version ($\textbf{Q}_{\overline{\beta}}$). The baseline is built on the shortest paths in the DBpedia-based multilingual graph.}
\label{tab:multilang}
\end{table*}

\begin{table*}
    \centering
    \begin{tabular}{|l|c|c|c|c|c|c|c|}

    \hline
  & \textbf{Baseline} & $\hat{\textbf{Q}}$ (\textit{avg}) &$\textbf{Q}$ (\textit{avg}) & $\textbf{Q}_{\overline{\beta}}$ (\textit{avg}) & $\hat{\textbf{Q}}$ (\textit{sif}) & $\textbf{Q}$ (\textit{sif}) & $\textbf{Q}_{\overline{\beta}}$ (\textit{sif}) \\ \hline
   \textbf{Lf} + \textbf{Tt} $\Longrightarrow$ \textbf{Dc} & $76.2 \pm 0.1$ & $75.2 \pm 0.2$& $82.0 \pm 0.2$ & $83.0 \pm 0.2$ & $81.3 \pm 0.2$ & $87.3 \pm 0.1$ & $\textbf{87.5} \pm \textbf{0.0}$\\ \hline
   \textbf{Dc} + \textbf{Tt} $\Longrightarrow$ \textbf{Lf} & $84.5 \pm 0.2$ & $81.6 \pm 0.2$ & $87.2 \pm 0.1$ & $88.0 \pm 0.1$ & $84.6 \pm 0.1$ & $90.1 \pm 0.1$ & $\textbf{90.4} \pm \textbf{0.1}$ \\ \hline
   \textbf{Lf} + \textbf{Dc} $\Longrightarrow$ \textbf{Tt} & $82.5 \pm 0.3$ &$82.2 \pm 0.3$ & $87.8 \pm 0.2$ & $88.1 \pm 0.2$ & $86.4 \pm 0.1$ & $\textbf{91.6} \pm \textbf{0.2}$ & $\textbf{91.8} \pm \textbf{0.2}$ \\ \hline
    \end{tabular}
\caption{Macro-AUC (\%) in English multi-source music genre translation with standard deviation computed over 4 folds. Results are shown for different embedding initialization (\textit{avg} and \textit{sif}), used directly ($\hat{\textbf{Q}}$) or retrofitted with the original retrofitting ($\textbf{Q}$) or with our version ($\textbf{Q}_{\overline{\beta}}$). The baseline consists in tag alignment against English DBpedia music genres \cite{Epure2019}.}
\label{tab:multisource}
\end{table*}

When comparing the initialization strategies, we can observe that directly using \textit{sif} embeddings in translation outperforms the baseline, while \textit{avg} yields lower AUC scores.
For all languages as targets, the \textit{sif} initialization is consistently more effective than the \textit{avg} initialization.
A significant difference between the two types of retrofitting applied to both initialization strategies exists, our version resulting in higher AUC scores. 
By differentiating between the two relation types, equivalence and other relatedness, the music genre embeddings appear to encode more accurately their relations within and across languages.  

\subsection{Results on Multi-Source Genre Translation}
\label{sec:results2}

In Table \ref{tab:multisource}, we present the results of the English-language multi-source music genre annotation. 
We re-compute the baseline \cite{Epure2019} results using Equation \ref{eq:transf}.
Compared to the previously reported AUC scores \cite{Epure2019}, the re-computed ones are higher showing that the modified translation scoring function does not disadvantage the knowledge-based music genre embeddings derived with the baseline.
In contrast to the baseline, our most effective method, using \textit{sif} initialization and our version of retrofitting, yields consistently higher AUC scores.
The increase in performance is of $11.3$ percentage points for \textbf{Dc} as target, $5.9$ points for \textbf{Lf} as target and $9.3$ points for \textbf{Tt} as target.

The \textit{sif} initialization of tag embeddings results in higher AUC scores than \textit{avg} both when the embeddings are used directly as they are or retrofitted, in particular, when \textbf{Dc} is target.
Also, let us notice that directly using the embeddings initialized with \textit{sif} leads to an increased performance compared to the baseline for \textbf{Dc} and \textbf{Tt}.
Retrofitting the embeddings significantly increases the AUC scores for all tag systems as targets.
Compared to the experiments reported in Section \ref{sec:results1}, this time, we observe only a marginal difference between the original retrofitting and our version.

Further, we give more details about the translations enabled by the baseline and our retrofitted \textit{sif} embeddings.
Often, we yield better music genre mappings (e.g. we map \textit{Discogs:uk garage} on \textit{Tagtraum:garagerock}, while the baseline maps it on \textit{Tagtraum:dubstep}).
However, there are also cases where the baseline leads to more accurate mappings (e.g. \textit{Discogs:modal} is mapped on \textit{Lastfm:cooljazz} compared to our best mapping on \textit{Lastfm:jazz}).
Finally, the baseline could not map at all some music genres, while we could (e.g. we map \textit{Discogs:crunk} on \textit{Tagtraum:gangstarap} and on \textit{Lastfm:rap}).

To sum up, exploiting the semantics of the music genre graph edges leads to marginally improved results w.r.t. the original retrofitting in the English-language multi-source translation and significantly higher AUC scores in the cross-lingual translation.
The \textit{sif} initialization yields better translations than the \textit{avg} initialization.
Lastly, we outperform the baselines by large margins in both experiments.

\section{Conclusion}
\label{sec:conclusion}
In this paper, we presented a new multi-step method for multilingual music genre embeddings learning. 
This method combines pre-trained word embeddings, music genre graphs and a retrofitting method leveraging different types of music genre relations to adapt embeddings to the music domain and learn embeddings for music genres with unknown words in the pre-trained word embeddings vocabulary.
Our experiments demonstrate the effectiveness of the proposed method, both in the English-language multi-source and the new cross-lingual translation tasks.

For future work, we plan to learn embeddings for each music genre relation type. 
Fang et al. \cite{Fang2019} consider that each edge represents a linear translation from the embedding of one node to the embeddings of its neighbour.
In a generalized setup, functional retrofitting proposed by Lengerich et al. \cite{lengerich2018} defines a linear relational penalty function for each type of relation in the graph.

Then, we aim to address the incremental updates of the music genre graph in order to avoid re-applying retrofitting every time the graph is updated.
Instead of relying on the constraints represented by the graph edges directly in retrofitting, Glavaš and Vulič \cite{glavas2018} use them as training instances to learn an explicit retrofitting function, which can be after applied to new node embeddings.

Further, we want to apply our method to new languages, especially from other language families, as well as to investigate other pre-trained word embeddings and alternatives to embed multi-word concepts \cite{shwartz-2019-systematic}.
For this, the current music genre graph needs to be populated with new multilingual music genres and their relations, and a parallel corpus of music items covering new languages should be collected if further evaluation is required. 
Continuing to rely on the multilingual DBpedia is an option, though a limiting one, given that only some world languages are supported.
Thus, music genre translation involving resource-poor languages remains a challenge. 
However, for the supported languages, our approach allows generating cross-lingual music genre annotations, which could be useful for other music information retrieval and recommendation tasks such as language-aware music genre auto-tagging, localized playlist captioning and music genre-driven recommendations, cross-cultural music genre perception modeling for user studies.  

Finally, the multilingual data 
and the code to learn and evaluate music genre embeddings are made available to the community\footnote{\href{https://github.com/deezer/MultilingualMusicGenreEmbedding}{https://github.com/deezer/MultilingualMusicGenreEmbedding}}.
Also, a demo to visualize the music genre vector space and the cross-lingual translation results for DBpedia music items is available \cite{epure2020muzeeglot}.

\section{ACKNOWLEDGEMENTS}
We would like to thank Manuel Moussallam, Marion Baranes, Anis Khlif and the ISMIR reviewers for their insightful and helpful comments on the paper.

\bibliography{refs}

\begin{thebibliography}{10}
\providecommand{\url}[1]{#1}
\csname url@samestyle\endcsname
\providecommand{\newblock}{\relax}
\providecommand{\bibinfo}[2]{#2}
\providecommand{\BIBentrySTDinterwordspacing}{\spaceskip=0pt\relax}
\providecommand{\BIBentryALTinterwordstretchfactor}{4}
\providecommand{\BIBentryALTinterwordspacing}{\spaceskip=\fontdimen2\font plus
\BIBentryALTinterwordstretchfactor\fontdimen3\font minus
  \fontdimen4\font\relax}
\providecommand{\BIBforeignlanguage}[2]{{%
\expandafter\ifx\csname l@#1\endcsname\relax
\typeout{** WARNING: IEEEtran.bst: No hyphenation pattern has been}%
\typeout{** loaded for the language `#1'. Using the pattern for}%
\typeout{** the default language instead.}%
\else
\language=\csname l@#1\endcsname
\fi
#2}}
\providecommand{\BIBdecl}{\relax}
\BIBdecl

\bibitem{Mandel2010}
M.~I. Mandel, D.~Eck, and Y.~Bengio, ``Learning tags that vary within a song,''
  in \emph{Conference of the International Society for Music Information
  Retrieval}, 2010.

\bibitem{Schedl2017}
M.~Schedl and B.~Ferwerda, ``Large-scale analysis of group-specific music genre
  taste from collaborative tags,'' in \emph{IEEE International Symposium on
  Multimedia}, 2017.

\bibitem{Craft}
A.~J. Craft, G.~A. Wiggins, and T.~Crawford, ``How many beans make five? {T}he
  consensus problem in music-genre classification and a new evaluation method
  for single-genre categorisation systems,'' in \emph{Conference of the
  International Society on Music Information Retrieval}, 2007.

\bibitem{Lee}
J.~H. Lee, K.~Choi, X.~Hu, and J.~Downie, ``K-pop genres: A cross-cultural
  exploration,'' in \emph{Conference of the International Society on Music
  Information Retrieval}, 2013.

\bibitem{Sordo2008}
M.~Sordo, O.~Celma, M.~Blech, and E.~Guaus, ``{The Quest for Musical Genres: Do
  the Experts and the Wisdom of Crowds Agree?}'' in \emph{Conference of the
  International Society on Music Information Retrieval}, 2008.

\bibitem{Epure2019}
E.~V. Epure, A.~Khlif, and R.~Hennequin, ``Leveraging knowledge bases and
  parallel annotations for music genre translation,'' in \emph{Conference of
  the International Society for Music Information Retrieval}, 2019.

\bibitem{Hennequin2018}
R.~Hennequin, J.~Royo-letelier, and M.~Moussallam, ``{Audio based
  disambiguation of music genre tags},'' in \emph{Conference of the
  International Society of Music Information Retrieval}, 2018.

\bibitem{mikolov2018}
T.~Mikolov, E.~Grave, P.~Bojanowski, C.~Puhrsch, and A.~Joulin, ``Advances in
  pre-training distributed word representations,'' in \emph{International
  Conference on Language Resources and Evaluation}, 2018.

\bibitem{pennington2014}
J.~Pennington, R.~Socher, and C.~D. Manning, ``{G}love: Global vectors for word
  representation,'' in \emph{Conference on Empirical Methods in Natural
  Language Processing}, 2014.

\bibitem{grave2018}
E.~Grave, P.~Bojanowski, P.~Gupta, A.~Joulin, and T.~Mikolov, ``Learning word
  vectors for 157 languages,'' in \emph{International Conference on Language
  Resources and Evaluation}, 2018.

\bibitem{artetxe2018}
M.~Artetxe, G.~Labaka, and E.~Agirre, ``A robust self-learning method for fully
  unsupervised cross-lingual mappings of word embeddings,'' in \emph{Annual
  Meeting of the Association for Computational Linguistics}, 2018.

\bibitem{mikolov2013exploiting}
T.~Mikolov, Q.~V. Le, and I.~Sutskever, ``Exploiting similarities among
  languages for machine translation,'' \emph{arXiv preprint arXiv:1309.4168},
  2013.

\bibitem{Auer_2007}
S.~Auer, C.~Bizer, G.~Kobilarov, J.~Lehmann, R.~Cyganiak, and Z.~Ives,
  ``Dbpedia: A nucleus for a web of open data,'' in \emph{The Semantic
  Web}.\hskip 1em plus 0.5em minus 0.4em\relax Springer, 2007, pp. 722--735.

\bibitem{faruqui2015}
M.~Faruqui, J.~Dodge, S.~K. Jauhar, C.~Dyer, E.~Hovy, and N.~A. Smith,
  ``Retrofitting word vectors to semantic lexicons,'' in \emph{Conference of
  the North American Chapter of the Association for Computational Linguistics:
  Human Language Technologies}, 2015.

\bibitem{Oramas2017}
S.~Oramas, O.~Nieto, F.~Barbieri, and X.~Serra, ``Multi-label music genre
  classification from audio, text and images using deep features,'' in
  \emph{Conference of the International Society on Music Information
  Retrieval}, 2017.

\bibitem{Coviello2011}
E.~Coviello, R.~Miotto, and G.~R. Lanckriet, ``Combining content-based
  auto-taggers with decision-fusion,'' in \emph{Conference of the International
  Society on Music Information Retrieval}, 2011.

\bibitem{Bogdanov2017}
D.~Bogdanov, A.~Porter, J.~Urbano, and H.~Schreiber, ``Mediaeval 2017
  acousticbrainz genre task: content-based music genre recognition from
  multiple sources,'' in \emph{MediaEval 2017 AcousticBrainz}, 2017.

\bibitem{schreiber2016}
H.~Schreiber, ``Genre ontology learning: Comparing curated with crowd-sourced
  ontologies.'' in \emph{Conference of the International Society for Music
  Information Retrieval}, 2019.

\bibitem{Achichi2018}
M.~Achichi, P.~Lisena, K.~Todorov, R.~Troncy, and J.~Delahousse, ``Doremus: A
  graph of linked musical works,'' in \emph{International Semantic Web
  Conference}, 2018.

\bibitem{Lisena2018}
P.~Lisena, K.~Todorov, C.~Cecconi, F.~Leresche, I.~Canno, F.~Puyrenier,
  M.~Voisin, T.~Le~Meur, and R.~Troncy, ``Controlled vocabularies for music
  metadata,'' in \emph{Conference of the International Society on Music
  Information Retrieval}, 2018.

\bibitem{smith2012distinct}
R.~D. Smith, ``Distinct word length frequencies: distributions and symbol
  entropies,'' \emph{Glottometrics}, vol.~23, pp. 7--22, 2012.

\bibitem{harris1954}
Z.~S. Harris, ``Distributional structure,'' \emph{Word}, vol.~10, no. 2-3, pp.
  146--162, 1954.

\bibitem{lijurafsky2015}
J.~Li and D.~Jurafsky, ``Do multi-sense embeddings improve natural language
  understanding?'' in \emph{Conference on Empirical Methods in Natural Language
  Processing}, 2015.

\bibitem{Gysel_2018}
C.~V. Gysel, M.~De~Rijke, and E.~Kanoulas, ``Neural vector spaces for
  unsupervised information retrieval,'' \emph{ACM Transactions on Information
  Systems (TOIS)}, vol.~36, no.~4, pp. 1--25, 2018.

\bibitem{joulin2018}
A.~Joulin, P.~Bojanowski, T.~Mikolov, H.~J{\'e}gou, and E.~Grave, ``Loss in
  translation: Learning bilingual word mapping with a retrieval criterion,'' in
  \emph{Conference on Empirical Methods in Natural Language Processing}, 2018.

\bibitem{shwartz-2019-systematic}
V.~Shwartz, ``A systematic comparison of {E}nglish noun compound
  representations,'' in \emph{Joint Workshop on Multiword Expressions and
  WordNet}.\hskip 1em plus 0.5em minus 0.4em\relax Association for
  Computational Linguistics, 2019, pp. 92--103.

\bibitem{mandelbrot1953informational}
B.~Mandelbrot, ``An informational theory of the statistical structure of
  language,'' \emph{Communication theory}, vol.~84, pp. 486--502, 1953.

\bibitem{Zipf1949}
G.~K. Zipf, \emph{Human behavior and the principle of least effort}.\hskip 1em
  plus 0.5em minus 0.4em\relax MA, Addison-Wesle: Cambridge, 1949.

\bibitem{arora2016simple}
S.~Arora, Y.~Liang, and T.~Ma, ``A simple but tough-to-beat baseline for
  sentence embeddings,'' in \emph{International Conference on Learning
  Representations}, 2017.

\bibitem{Golub:1970}
G.~H. Golub and C.~Reinsch, ``Singular value decomposition and least squares
  solutions,'' in \emph{Linear Algebra}.\hskip 1em plus 0.5em minus 0.4em\relax
  Springer, 1971, pp. 134--151.

\bibitem{Miller_1995}
G.~A. Miller, ``Wordnet: a lexical database for english,'' \emph{Communications
  of the ACM}, vol.~38, no.~11, pp. 39--41, 1995.

\bibitem{lengerich2018}
B.~J. Lengerich, A.~L. Maas, and C.~Potts, ``Retrofitting distributional
  embeddings to knowledge graphs with functional relations,'' in
  \emph{International Conference on Computational Linguistics}, 2018.

\bibitem{Fang2019}
L.~Fang, Y.~Luo, K.~Feng, K.~Zhao, and A.~Hu, ``Knowledge-enhanced ensemble
  learning for word embeddings,'' in \emph{World Wide Web Conference}, 2019.

\bibitem{scheepers2018}
T.~Scheepers, E.~Kanoulas, and E.~Gavves, ``Improving word embedding
  compositionality using lexicographic definitions,'' in \emph{World Wide Web
  Conference}, 2018.

\bibitem{saad2003iterative}
Y.~Saad, \emph{Iterative methods for sparse linear systems}.\hskip 1em plus
  0.5em minus 0.4em\relax SIAM, 2003.

\bibitem{bengio2006label}
Y.~Bengio, O.~Delalleau, and N.~Le~Roux, ``Label propagation and quadratic
  criterion,'' \emph{Semi-Supervised Learning}, pp. 193--216, 2006.

\bibitem{saha2016dis}
T.~K. Saha, S.~Joty, N.~Hassan, and M.~A. Hasan, ``Dis-s2v: Discourse informed
  sen2vec,'' \emph{arXiv preprint arXiv:1610.08078}, 2016.

\bibitem{hayes2019}
D.~Hayes, ``What just happened? {E}valuating retrofitted distributional word
  vectors,'' in \emph{Conference of the North {A}merican Chapter of the
  Association for Computational Linguistics: Human Language Technologies},
  2019.

\bibitem{speer2016ensemble}
R.~Speer and J.~Chin, ``An ensemble method to produce high-quality word
  embeddings,'' \emph{arXiv preprint arXiv:1604.01692}, 2016.

\bibitem{Ibrahim2020}
K.~Ibrahim, J.~Royo-Letelier, E.~Epure, G.~Peeters, and G.~Richard,
  ``Audio-based auto-tagging with contextual tags for music,'' in
  \emph{International Conference on Acoustics, Speech, and Signal Processing},
  ser. ICASSP, 05 2020, pp. 16--20.

\bibitem{Sechidis:2011}
K.~Sechidis, G.~Tsoumakas, and I.~Vlahavas, ``On the stratification of
  multi-label data,'' in \emph{Joint European Conference on Machine Learning
  and Knowledge Discovery in Databases}, 2011.

\bibitem{joulin2017}
A.~Joulin, E.~Grave, P.~Bojanowski, and T.~Mikolov, ``Bag of tricks for
  efficient text classification,'' in \emph{Conference of the {E}uropean
  Chapter of the Association for Computational Linguistics}, 2017.

\bibitem{glavas2018}
G.~Glava{\v{s}} and I.~Vuli{\'c}, ``Explicit retrofitting of distributional
  word vectors,'' in \emph{Annual Meeting of the Association for Computational
  Linguistics}, 2018.

\bibitem{epure2020muzeeglot}
E.~V. Epure, G.~Salha, F.~Voituret, M.~Baranes, and R.~Hennequin, ``Muzeeglot:
  annotation multilingue et multi-sources d'entit{\'e}s musicales {\`a} partir
  de repr{\'e}sentations de genres musicaux,'' in \emph{6e conf{\'e}rence
  conjointe Journ{\'e}es d'{\'E}tudes sur la Parole (JEP, 31e {\'e}dition),
  Traitement Automatique des Langues Naturelles (TALN, 27e {\'e}dition),
  Rencontre des {\'E}tudiants Chercheurs en Informatique pour le Traitement
  Automatique des Langues (R{\'E}CITAL, 22e {\'e}dition). Volume 4:
  D{\'e}monstrations et r{\'e}sum{\'e}s d'articles internationaux}.\hskip 1em
  plus 0.5em minus 0.4em\relax ATALA, 2020, pp. 18--21.

\end{thebibliography}

\end{document}